\documentclass{article} 

\usepackage{arxiv}

\usepackage{natbib}

\usepackage{amsmath,amsfonts,bm}

\def\eqref#1{equation~\ref{#1}}

\def\1{\bm{1}}

\DeclareMathAlphabet{\mathsfit}{\encodingdefault}{\sfdefault}{m}{sl}
\SetMathAlphabet{\mathsfit}{bold}{\encodingdefault}{\sfdefault}{bx}{n}

\DeclareMathOperator*{\argmax}{arg\,max}

\usepackage{hyperref}
\hypersetup{
    colorlinks=true,
    linkcolor=red,
    filecolor=magenta,      
    urlcolor=blue,
    citecolor=purple,
    pdftitle={Overleaf Example},
    pdfpagemode=FullScreen,
    }

\usepackage{graphicx}
\usepackage{algorithm}
\usepackage{algpseudocode}
\usepackage{svg}

\usepackage{xcolor}

\title{A model-based approach to meta-Reinforcement Learning: Transformers and tree search}

\author{Brieuc Pinon, Jean-Charles Delvenne, Raphaël Jungers\\
ICTEAM/INMA\\
UCLouvain\\
Belgium \\
\texttt{\{brieuc.pinon,jean-charles.delvenne,raphael.jungers\}@uclouvain.be}
}

\date{August, 2022}

\begin{document}

\maketitle

\begin{abstract}
    Meta-learning is a line of research that develops the ability to leverage past experiences to efficiently solve new learning problems. Meta-Reinforcement Learning (meta-RL) methods demonstrate a capability to learn behaviors that efficiently acquire and exploit information in several meta-RL problems.
    
    In this context, the Alchemy benchmark has been proposed by \citet{wang2021alchemy}. Alchemy features a rich structured latent space that is challenging for state-of-the-art model-free RL methods. These methods fail to learn to properly explore then exploit.
    
    We develop a model-based algorithm. We train a model whose principal block is a Transformer Encoder to fit the symbolic Alchemy environment dynamics. Then we define an online planner with the learned model using a tree search method. This algorithm significantly outperforms previously applied model-free RL methods on the symbolic Alchemy problem.
    
    Our results reveal the relevance of model-based approaches with online planning to perform exploration and exploitation successfully in meta-RL. Moreover, we show the efficiency of the Transformer architecture to learn complex dynamics that arise from latent spaces present in meta-RL problems.

    
    
    
\end{abstract}


\section{Introduction}
    Deep Learning methods have been successfully applied to various problems such as in image processing, natural language processing, and games  \citep{schmidhuber2015deep,goodfellow2016deep,li2018deep}. However, these solutions usually require many samples to solve any new task. The quickly developing field of meta-Learning addresses this issue. To solve a task of interest, the meta-Learning paradigm supposes access to data from other tasks in relation to the task of interest. A meta-learning method can then take advantage of this supplementary data to learn an efficient learning algorithm for the task of interest.
    
    
    The meta-Learning literature has quickly broadened. Meta-learning, or learning to learn, has two learning levels where design decisions can be made by researchers. For example at the meta-learning level, different optimizers can be choosed such as gradient descent, Reinforcement Learning (RL) or evolution. We refer to \cite{hospedales2020meta} for a review of such designs. Furthermore, several types of learning problems have been studied: supervised, unsupervised, and RL. Moreover, different constraints on the relationships between the tasks might be supposed. In this paper, we work with an RL problem where we suppose that the tasks are sampled from an unknown but fixed probability distribution.
    
    
    Recently proposed by \citet{wang2021alchemy}, the Alchemy benchmark aims to test and understand proposed meta-RL algorithms. It poses the problem with a fixed distribution over the space of tasks. Each of these tasks corresponds to a set of causal links between ``potions'' use and effects on ``stones''. These links are hidden from the agent. To solve the task, the agent must be able to experiment with potions in order to discover those links, then exploit that knowledge to maximize its rewards. This implicitly requires that the agent understands the structure of the latent dynamics. Two versions of the benchmark have been proposed, a 3D visual version and a symbolic version. We focus on the symbolic version in this paper.
    
    One approach to meta-RL is to cast the problem as a partially observable Markov Decision Process (POMDP), where a latent space represents the hidden dynamics. Model-free RL methods that support partial observability can then be used for meta-RL \citep{wang2016learning,duan2016rl}. The authors of Alchemy tested state-of-the-art model-free RL methods on it. Their experimentation revealed a failure of these model-free RL methods to learn a policy that efficiently explores for information and then exploit it both on the 3D visual and on the symbolic case \citep{wang2021alchemy}. 
    
    We investigate the use of a model-based algorithm with online planning on the symbolic version of Alchemy and show significant improvements.
    This result shows both:
    \begin{itemize}
        \item the capability of Deep Learning methods, in particular the Transformer architecture, to fit complex dynamics in environments where model-free RL methods fail. These dynamics can emerge in meta-RL problems where the latent space is a critical part;
        \item the strength of online planning algorithms in challenging environments, such as those that arise in meta-RL where the reward is delayed between the gain in information from exploration and its exploitation.
    \end{itemize}
    

    \subsection{Related literature} 
        
        We survey model-free RL applied to meta-RL problems literature since it is the solution studied by \cite{wang2021alchemy}. We then describe model-based approaches broadly and then focus on model-based RL applied in meta-RL. We also mention other solutions that use models of the dynamics in meta-RL. 
        
        \subsubsection{Model-free RL for meta-RL}
            As said, one approach to meta-RL is to pose the construction of the efficient agent as an RL problem with partial observability. Then this POMDP can be solved with classical deep RL methods. A usual way in deep RL to support partial observability is to make the past observations input to the neural networks representing the policy or (Q-)value function. Several architectures that support these sequential inputs have been applied: long short-term memory (LSTM) and gated recurrent units (GRU) \citep{wang2016learning,duan2016rl}, memory-augmented neural-networks by \citep{santoro2016meta}, a mix of temporal convolution and attention \citep{mishra2017simple}, and an architecture based on Transformers \citep{parisotto2020stabilizing}.
        
            Several works use a model-free RL algorithm and complement it with auxiliary losses that force an inner representation of the studied system dynamics \citep{wayne2018unsupervised,gelada2019deepmdp,zintgraf2019varibad}. They use modeling losses to have self-supervision and create potentially relevant abstractions, but do not use their ability to predict the dynamics of the system directly for planning. Similarly, a task description can be used as a target to construct useful abstractions \citep{humplik2019meta}. It has been proposed that model-free methods in meta-RL implicitly learn to do such task inference \citep{alver2021going}. 

        \subsubsection{Model-based approaches}
            In model-based approaches, a model of the dynamics is learned. Then the model can be used inside a planner to decide on actions to take inside the environment. Another possibility is to learn a policy or (Q-)value function from the learned model. This can be done in several ways: using the capacity to simulate trajectories and so get free samples; using the potential differentiability of the learned model to compute gradients for the agent through the simulated steps by the model \citep{heess2015learning}; using a planner to create target behavior for the agent, then using the improved agent to improve the planner, etc. in a virtuous cycle \citep{schrittwieser2020mastering}.
            
            Another design decision of model-based RL methods is the class of functions used to fit the dynamics, such as linear models or neural networks. Similarly to model-free methods, the support for partial observability allows applying model-based RL methods on meta-RL problems.
            
            Efficient methods based on Gaussian Processes have been developed \citep{deisenroth2011pilco}. These methods can be extended to support partial observability and thus meta-RL problems \citep{saemundsson2018meta}. However, Gaussian Processes algorithms rely on kernel methods for generalization and thus inherit their limitations.
            
            
            In the context of meta-RL with neural networks: \citet{hiraoka2021meta} train a policy from simulated trajectories and provide theoretical arguments for their method; \citet{perez2020generalized} uses probabilistic inference on latent variables representing the hidden dynamics, and combine it with the cross-entropy method for planning.
        
            We mention the theoretical work by \citet{sun2019model}, which analyzes the potential gap in the number of samples required to reach some performance between model-free and model-based methods. The argumentation shows that a significant gap can exist to the advantage of model-based methods. To prove the gap, they construct a family of systems whose dynamics factorize, i.e. the evolution of each variable only depends on a subset of all the variables. The Alchemy environment has precisely a factorizable dynamics. So, the results in our paper go in the same direction as the highlighted sample complexity gap and its argumentation.
        
        \subsubsection{Fast model-adaptation}
            We presented solutions where the model was fitted on the dynamics comprising the latent space that defined the current dynamic of the episode. Another way to make use of model-based RL in a meta-learning context is to have a fixed model across all tasks/episodes that does not fit the latent dynamic. Instead, its weights are trained to quickly adapt to the current dynamics with gradient steps. This is an application of the Model-Agnostic Meta-Learning (MAML) method \citep{finn2017model} on the model \citep{nagabandi2018learning,nagabandi2018deep,mendonca2020meta}.
        
        \subsubsection{Fixed model}
            In the meta-RL literature, a component of the model can be held fixed, for example a fixed state to state dynamics but with a non-fixed reward function. However, this can only work for problems where the set of tasks respect strong constraints on their dynamics. An example of such a work is \citet{byravan2020imagined}.
        
        
        \subsubsection{Our solution}
            In relation to the presented literature, the solution we define in the following section is a model-based algorithm that learns a model of the dynamics with a Transformer Encoder and then incorporates it in a tree search planner without learning any policy or value functions.
            
            This solution can be incorporated in an RL algorithm that learns a neural network policy/value function using target behaviors produced by our algorithm. The policy and value function can then be used to improve the planner in a virtuous cycle, like in the cited work by \cite{schrittwieser2020mastering}.

\section{Methods}
    We define the general problem we address \ref{sec:problem-form}, and describe the symbolic Alchemy benchmark \ref{sec:symb-alchemy}. We then explain the construction of our model-based agent. How we learn a neural network model of the dynamics \ref{sec:learn-model}, and what online planner we use on top of this model \ref{sec:tree-search}.
    
    \subsection{Problem formulation}\label{sec:problem-form}
        
        We define a partially observable Markov decision process (POMDP) as a tuple $(S,A,T,P,\Omega,O,N)$ where $S$ is the set of states, $A$ is the set of actions, $T$ is the conditional transition probability distribution over states and rewards $T(s_{t+1},r_{t+1}|s_t,a_t)$ where $s_{t+1},s_t\in S$, $a_t\in A$ and $r_{t+1}\in \mathbb{R}$, and $P$ is the initial probability distribution over the states. The set $\Omega$ define the possible observations and $O$ is a conditional probability distribution $O(o_t|s_t,a_t)$ describing the link between the state and the observations given to the agent. The number $N$ gives the number of steps in one episode.
        
        In the meta-problem, at the start of each episode a POMDP $m$ is sampled from an unknown but fixed distribution over a set of POMDPs that share the same number of steps by episode $N$, and observation and action spaces $\Omega$ and $A$. We then undergo the dynamics, rewards and observations of $m$ until the end of the episode.
        
        Our goal is to define an agent (player) that acts in the meta-problem and maximizes its expected sum of rewards.
    
    \subsection{Symbolic Alchemy}\label{sec:symb-alchemy}
        Symbolic Alchemy by \citet{wang2021alchemy} links a ``chemistry'' to the dynamics of a (PO)MDP. At each episode a chemistry/MDP is sampled. The goal in the episode is to explore and use to our advantage acquired knowledge about this chemistry. We define next what constitutes a state, the action space and the unrolling of one episode.
        
        
        In symbolic Alchemy, there are three types of objects that determine a state: the chemistry; the stones, their respective 3 visual features and associated reward; and the potions with their respective colors. The stones, potions, and associated features are directly observable, the chemistry is not.
        
        The space of actions is: applying a potion on a stone, this can change its perceptual features and reward; collecting the reward associated with a stone (can only be done one time for each stone); and a no-operation. The hidden chemistry determines the effects of potions on stones and the possible features of stones.
        
\begin{figure}
            \centering
            \includegraphics[width=0.4\linewidth]{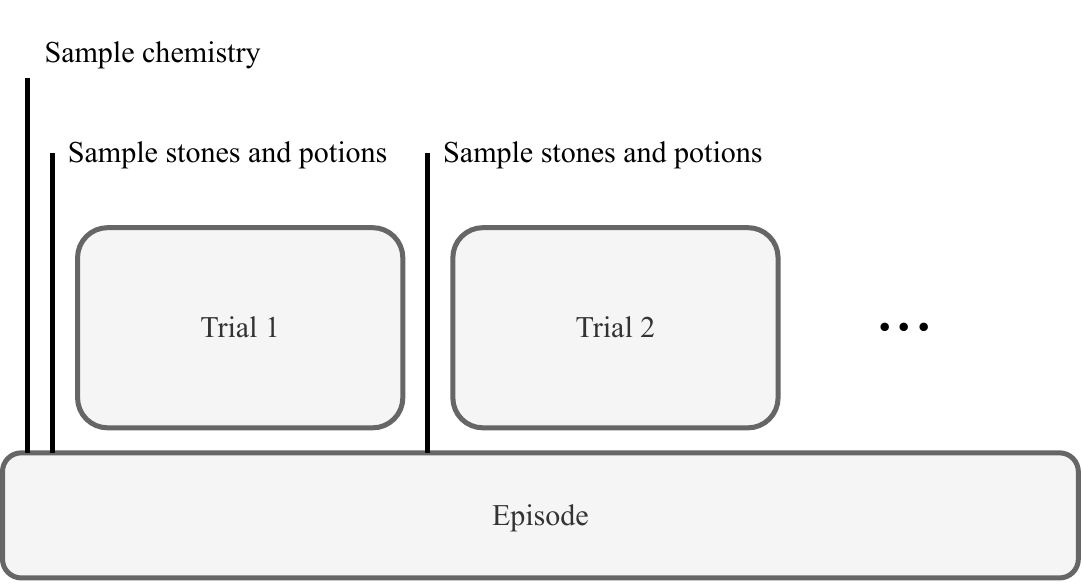}
            \caption{Decomposition of one episode of symbolic Alchemy: a chemistry is sampled for the episode, then trials are interleaved with resamples of new potions and stones.}
            \label{fig:symbolic-alchemy_episode}
        \end{figure}

        An episode is composed as follows: a chemistry is sampled at the start and is kept constant during the whole episode; then a sequence of 10 trials follows. Each trial is further decomposed in: sampling a set of stones and potions; then 20 time-steps where actions can be taken to understand the chemistry through experiments, generating high-value stones by using potions, and collecting rewards associated to the stones. The decomposition of one episode is given on Figure \ref{fig:symbolic-alchemy_episode}.
        
        The sampled chemistry respects several constraints:
        \begin{itemize}
            \item The allowed feature values of stones are determined by the rotation of a cube in 3D. The reward of a stone only depends on its perceptual features. So, the states of the stones are uniquely determined by their positions on the vertices of a 3D cube.
            \item The potion color and stone position uniquely determine the transition of the stone position when the potion is applied. Those transitions only happen between vertices linked by an edge in the underlying cube.
            \item Moreover, the effect of a potion with a specific color happens upon parallel edges of the cube and in the same direction.
            \item Potion colors come in pairs. If one potion color produces transitions in one direction of the stone positions, then its pair does the same in the opposite direction. The pairs are consistent across episodes.
            \item Some transition/edges of the cube can be missing. The topology of the existing edges respect constraints, for example, the connectivity of the cube must hold.
        \end{itemize}
        
        For an agent, all of these constraints are opportunities to quickly discard causal links between potions applications and stones features changes. Thus, an optimal agent must understand and use these constraints.

    \subsection{Learning a model}\label{sec:learn-model}
        \subsubsection{Data}
            We sample a dataset of trajectories prior to any learning and planning. We simply use a uniform policy over the action space to generate the trajectories. We define a supervised learning problem from this data.
            
            While this method is sufficient to learn an accurate model in the case of symbolic Alchemy, other environments might need interleaving data generation and model learning to get a sufficient coverage of the state and action spaces.
        
        \subsubsection{Model architecture}
\begin{figure}[ht]
                \begin{center}
                \includegraphics[width=0.75\linewidth]{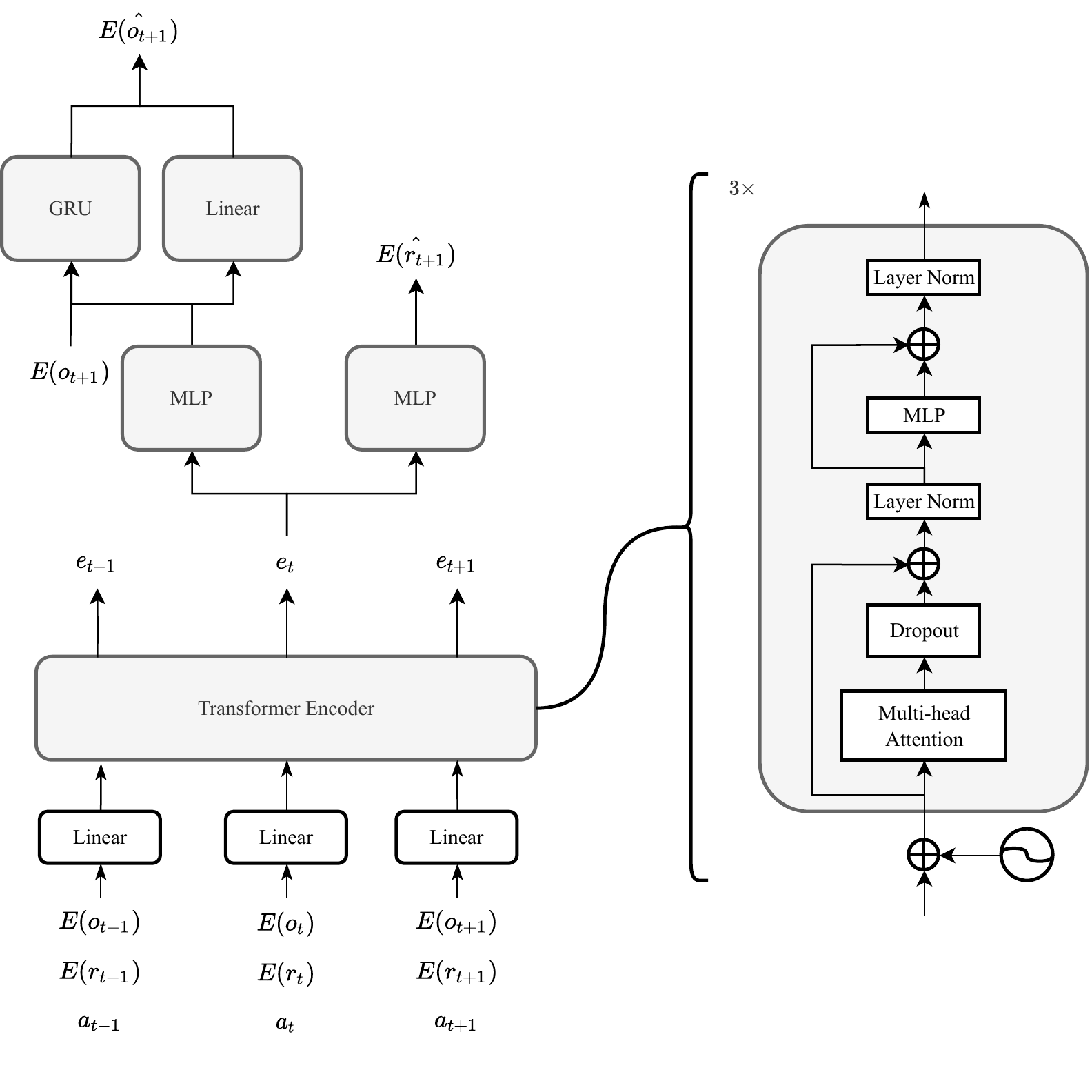}
                \caption{A neural network architecture to fit trajectories. The sequence of observations and rewards are first transformed into categorical variables. Then the sequence is passed through a linear layer along with the sequence of actions. The resulting sequence is fed to a Transformer Encoder with three layers, causal attention, and a sinusoidal positional encoding. Encodings, $e_t$, from the Transformer Encoder are then used independently to predict the next elements in the sequences. For the rewards, an MLP is used. For the observations, first an MLP then a GRU plus a linear layer are combined. We note that the next observation is also in input for the GRU since we use teacher forcing.}
                \label{fig:model_arch}
                \end{center}
            \end{figure}

            The architecture is presented in Figure \ref{fig:model_arch}. Our model is based on a Transformer Encoder using causal attention for its main block \citep{vaswani2017attention}. It receives as input a sequence of observations, rewards, and actions. And it outputs a prediction for the next observation and reward at each time step.
            
            We use the Transformer architecture to fit the dynamics for its capacity to learn complex dynamics, as demontrated in natural language processing.
            
            In the prediction head for the next observation, the GRU allows our neural network to model non-independent probability distributions over the dimensions of the next observation. In other words, our model is not constrained to represent distributions $p(o_{t+1})$ that factorize in $\Pi_i p_i(o_{t+1}^i)$ where $i$ is the $i$th dimension of the observation. A more detailed view of this part of the model is available in Figure \ref{fig:teacher-forcing} in Appendix \ref{sec:model-details}. We show in our results that both the linear and GRU parts are necessary to have an accurate model of the dynamics of symbolic Alchemy.
            
            To train our model, we use teacher forcing. At training time to predict the $i$th dimension of the observation at timestep $t+1$, we use the true observations and rewards up to the timestep $t$, and also, the ground truth up to the $i-1$th dimension of the observation at timestep $t+1$, see Figure \ref{fig:teacher-forcing} in Appendix \ref{sec:model-details} for a detailed representation of teacher forcing.
            
            For the specific case of symbolic Alchemy, we encode the real-valued observations and rewards as categorical variables both for the input and output. Discretizing the inputs allows us to have efficient cross-entropy loss and helps to get consistent trajectory simulation over longer time horizon. We note that this is not unusual in the literature, for example $\mu$-zero discretizes the prediction of the reward \cite{schrittwieser2020mastering}. The encoding, $E$, is a one-hot transformation that takes as input a $d$-dimensional real-valued vector and returns a $d\times n$-tensor where $n$ is the number of categories. The number $n$ and the values used in the encoding can be extracted from the generated offline data to cover all the possibilities for symbolic Alchemy.
            
            A full description of the hyperparameters that we use for the Alchemy benchmark can be found in the Appendix \ref{sec:hyper}.
            
    \subsection{Tree search with stochastic transitions}\label{sec:tree-search}
        We define the model-based online planning algorithm we use. Note that this planning algorithm uses for its simulations the learned model defined in the previous section.
    
        We construct our online planner on top of the tree search planner presented in \cite{silver2016mastering}, and more specifically, the version defined by \cite{schrittwieser2020mastering}. The tree search algorithm grows iteratively a tree where each branch represents a potential sequence of actions starting at the current state. The original algorithm supposes full observability, deterministic transitions, a policy and a value function. We adapt it to support partial observability and stochastic transitions.
        
        We use a common transformation that casts POMDPs as equivalent MDPs where states correspond to the complete history of observations up to that point, see for example \cite{baxter2000direct}. The transformation ensures that a policy defined on the MDP gives a behavior with the same performance on the original POMDP. Note that even if the original POMDP was deterministic, the resulting MDP is stochastic in general. 
        
        To support stochastic dynamics, we adapt the tree search algorithm by adding chance nodes as children of decision nodes. At chance nodes, we branch different paths representing different stochastic realizations of the system dynamics. For each chance node, we use a fixed number of samples from which stochastic branches are created. We merge identical samples to reduce the tree breadth. When we merge samples, we use weights to not introduce any bias.
        
        
        To compensate for our lack of policy and value function, we use defaults: a uniform probability distribution over actions; and a zero-everywhere value function is taken.
        
        The algorithm grows the tree by prioritizing more promising branches. At each node, it uses Q-value estimates and also uncertainty estimates on these values to quantify how interesting an action might be.
        
        The algorithm cycles through 3 steps to grow the three:
        \begin{itemize}
            \item \textit{Selection} Starting at the root node, decision nodes' branches are taken depending on the Q-value estimates, the uncertainty on these estimates, and the size of the subtree at that node. At chance nodes, we sample a branch with a probability proportional to its weight. We stop when we have no child for the action to take.
            \item \textit{Expansion} Where we block in the selection step, we create a chance node and sample its children.
            \item \textit{Backup} With the new rewards obtained at the expansion step, we update the Q-value estimates on the decision nodes taken in the selection step.
        \end{itemize}
        
\begin{figure}[ht]
            \begin{center}
            \includegraphics[width=0.7\linewidth]{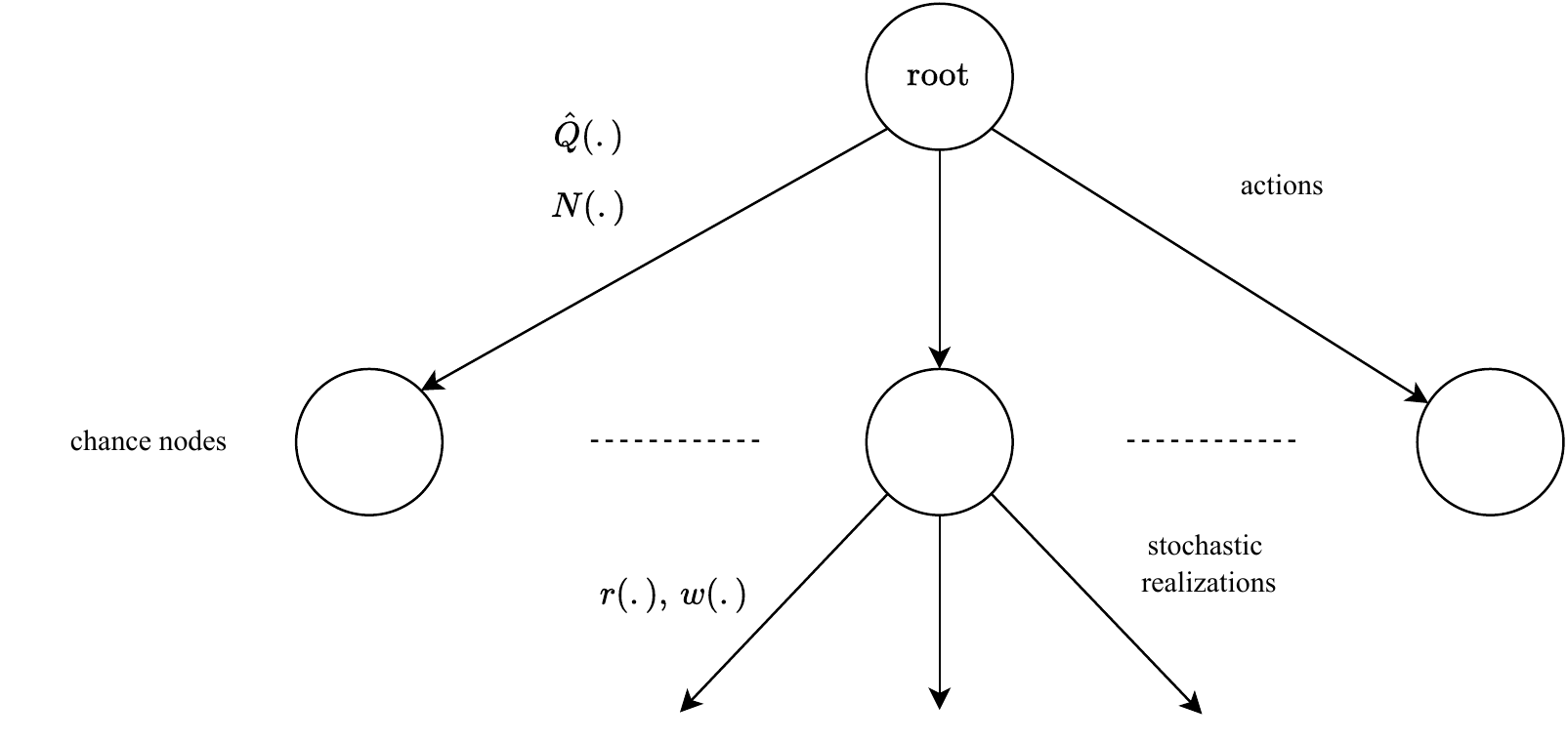}
            \caption{Tree search algorithm with support for stochastic transitions.}
            \label{fig:tree-search}
            \end{center}
        \end{figure}

        See Algorithm 1 for a more detailed description of the tree search and Figure \ref{fig:tree-search} for a partial representation of a tree.
        
        We note that since the number of samples by chance node is constant, a limited number of possible paths for the system dynamics are considered. Thus, in general, the Q-values estimates in the tree will not asymptotically converge to their true values with the size of the tree. However, in practice, the size of the tree is bounded and the estimates will have non-zero variance in any case. The number of samples by chance node is a hyper-parameter that can be optimized to compromise between the tree width and the lack of precision of these estimates.

        \begin{algorithm}
            \caption{Online tree search planning adapted from \cite{schrittwieser2020mastering}}
            \begin{algorithmic}
                \label{alg:tree-search}
                \Require a learned model; a discount factor $\gamma$; the number of sampled stochastic branches by chance nodes $K$; exploration parameters for the selection $C_1,C_2$; temperature parameter for action probabilities $T$; a number of expansions to run.
                \State
                \State initialize a root node with the current state
                \ForAll {1\ldots\# expansions}
                \State
                    \State \textbf{Select}
                    \State decision node $\leftarrow$ root node
                    \Loop
                        \State $a^* \leftarrow \argmax_{a\in A} Q^{\text{node}}(a) + \frac{1}{|A|}\frac{\sqrt{\sum_a N^{\text{node}}(a)}}{1+N^{\text{node}}(a)}(C_1+\log(\frac{C_2+1+\sum_a N^{\text{node}}(a)}{C_2})) $
                        \State chance node $\leftarrow$ take child of decision node following $a^*$ \textbf{if} it does not exist \textbf{break}
                        \State decision node $\leftarrow$ sample child of chance node
                    \EndLoop
                    \State 
                    \State \textbf{Expand}
                    \ForAll{1\ldots K}
                        \State sample $s,r\leftarrow$ learned model(decision node state,$\,a^*$)                    
                    \EndFor
                    \State compare samples, merge duplicates and increase the sampling weight $w$
                    \State create chance node and its children decision nodes with sampled states
                    \State assign new child chance node to the decision node at the end of select following $a^*$
                    \State
                    \State \textbf{Backup}
                    \State $G\leftarrow 0.$
                    \State $r\leftarrow \sum_{(s,r,w)\in\text{branches of chance node}} w\cdot r / \sum_{(s',r',w')\in\text{branches of chance node}}w' $
                    \State $a\leftarrow a^*$
                    \Repeat
                        \State $G\leftarrow r+\gamma G$
                        \State $Q^{\text{decision node}}(a) \leftarrow \frac{Q^{\text{decision node}}(a)\cdot N^{\text{decision node}}(a)+G}{ N^{\text{node}}(a)+1}$
                        \State $N^{\text{decision node}}(a) \leftarrow  N^{\text{decision node}}(a)+1$
                        \State chance node, $r$ $\leftarrow$ parent of decision node and associated reward
                        \State decision node, $a$ $\leftarrow$ parent of chance node and associated action
                    \Until {root node is updated}
                \EndFor
                \State
                \State \textbf{Actions probabilities}
                \ForAll{$a\in A$}
                    \State action probability of $a$ $\leftarrow N^{\text{root}}(a)^{1/T}/\sum_{a'\in A}N^{\text{root}}(a')^{1/T}$
                \EndFor
            \end{algorithmic}
        \end{algorithm}
        
\section{Results}
    We compare our method against the results obtained by \citet{wang2021alchemy}. They tested V-MPO a state-of-the-art model-free RL method \citep{song2019v}, applied with a Transformer architecture improved for RL \citep{parisotto2020stabilizing}. They also implemented an ideal observer, which is a Bayes-optimal agent that has the best performance achievable on the benchmark. This agent is not directly comparable to the others, since its implementation relies on prior knowledge of the problem. For further insights into the agent performances, they also implemented a simple Random Heuristic which does not exploit information efficiently.
    
    We sample $1e6$ episodes to construct our dataset of trajectories from which we learn our model. In the paper \cite{wang2021alchemy}, the authors report that $1e9$ episodes were used to train the V-MPO agent on symbolic Alchemy.
    
    See Table \ref{comparison-table} for the comparison, we test our method with several magnitudes of computational resources going into planning. We optimize the hyperparameters of the tree-search for the case of 1.250 expansions by step in the environment, see Appendix \ref{sec:hyper} for the values.
    
    The comparison reveals that our method surpasses the model-free RL method V-MPO if given sufficient computational resources for online planning. It also shows the importance of planning on this benchmark, the performance increases significantly with more allocation of resources into planning. However, the marginal gains diminish too fast to reach the optimal performance in our tests.
    
    \begin{table}[h]
    \caption{Comparison with the V-MPO RL method and the baselines given in \citet{wang2021alchemy}. We give the scores with the standard error. The number of expansions refers to the expansion operation defined in the description of the tree search.}
    \label{comparison-table}
    \begin{center}
    \begin{tabular}{ll}
        \multicolumn{1}{c}{\bf Agent}  &\multicolumn{1}{c}{\bf Score on symbolic Alchemy}
        \\ \hline \\
        Ours (\# expansions 100)        & $79.3\pm1.6$\\
        Ours (\# expansions 500)        & $161.8\pm3.9$\\
        Ours (\# expansions 1.250)        & $207.1\pm3.5$\\
        Ours (\# expansions 2.500)    & $220.5\pm3.2$\\
        Ours (\# expansions 5.000)        & $229.5\pm3.9$\\ 
        Ours (\# expansions 10.000)        & $251.5\pm4.5$\\ 
        V-MPO         & $155.4\pm1.6$ \\\hline
        Ideal Observer             & $284.4\pm1.6$ \\
        Random Heuristic             & $145.7\pm1.5$ \\\hline
    \end{tabular}
    \end{center}
    \end{table}
    
    \subsection{Architecture tests}
        We conduct an ablation study on our model's architecture to identify the essential elements to its success. We do three tests: replacing the Transformer Encoder with a recurrent architecture, a GRU; suppressing the GRU in the prediction of the next observation; and suppressing the linear layer in the prediction of the next observation. We test the capacity of the architecture to fit the dynamics of the symbolic Alchemy environment. We use this metric instead of the direct score because it better measures the potential of the learned model to construct an optimal policy. An error in the learned model induces a limit on the performance of the behavior for any online or offline planning method that we would build with the learned model. Results are given in Table \ref{ablation-table}.
        
        \begin{table}[h]
            \caption{All other hyperparameters being equal to the original architecture, see \ref{sec:hyper}. With an exception when we remove the final linear layer, where to increase performance, we decrease the learning rate to $2e-5$ and we double the number of hidden units in the GRU to $64$. When we replace the Transformer with a GRU, the GRU has 3 layers each of dimension $512$. The entropy loss is computed on a held-out fraction of the generated dataset described in Section \ref{sec:learn-model}.}
            \label{ablation-table}
            \begin{center}
            \begin{tabular}{ll}
                \multicolumn{1}{c}{\bf Model architecture}  &\multicolumn{1}{c}{\bf entropy loss on validation set}
                \\ \hline \\
                original architecture & 1.31\\
                replace Transformer with GRU & 1.37\\
                without the GRU head & 1.61\\
                without the final linear layer & 2.25
            \end{tabular}
            \end{center}
        \end{table}
        
        If we replace the Transformer Encoder with a GRU, the prediction performance drops. In symbolic Alchemy, the entropy in the prediction has two sources: the uncertainty that comes from predicting the next state of a stone after applying a potion; and predicting the uncertainty at each trial start when the set of stones and potion is reinitialized. To further understand the difference in entropy loss, we measure the part of the entropy loss coming from the former which correspond to the uncertainty on the current chemistry. We obtain $0.066$ with the original model and $0.11$ with the GRU as a main component. So part of the structure in the symbolic Alchemy problem is harder to learn for the GRU architecture, where in comparison, the Transformer architecture successfully grasps the latent dynamics.
        
        When we remove the GRU head in the prediction of the next observation, we decrease the model capacity since we forbid the output to represent a probability distribution of independent variables (one variable by feature of the observation). The gap in predictive performance with the original architecture shows the importance of modeling the interdependence between the different variables in an observation.
        
        We observe that without the final linear layer to help predict the next observation, optimizing the model is much harder. We suppose that this is not due to a lesser capacity of the model, since we increased the capacity of the GRU head to compensate for this effect. So we suspect that backtracking only through the GRU steps must be harder for the optimizer.

\section{Discussion}
    We demonstrate that a model-based approach with online planning can substantially surpass the performance of a state-of-the-art model-free method on the symbolic Alchemy meta-RL benchmark. We determine that this success relies on several factors: the use of the Transformer architecture for the model in comparison to a recurrent architecture; the use of a GRU to predict at a dimension-level the next-observation, so that our architecture is able to model non-independent probability distributions, and the use of a direct linear layer to improve optimization; an online planning algorithm allowed to use many simulated steps.
    
    Our work shows that while most of the research effort has focused on model-free methods in meta-RL \citep{hospedales2020meta}, learning to learn might need explicit modeling and then planning. Also, the use of the Transformer architecture can be critical to model the complex relationship between histories of observations and beliefs upon the system transitions.

    \paragraph{Improving the behavior} The online planner that we use can take as input a learned policy and value function to improve its performance. Moreover, a performant policy and value function can be learned from the online planner. Due to limited computational resources, we did not investigate this solution. We believe that this method would significantly improve the behavior and decrease the computational burden when we apply the policy, see \cite{hamrick2020role} for an empirical investigation of this method. This improvement would not increase the number of ground-truth samples required, but would significantly increase the computational requirements to train the agent.
    
    \paragraph{Abstract state-space} We only investigated a symbolic benchmark. To extend our solution to visual domains for example, we expect that modeling and planning should be made in an abstract state and not at the pixel level.
    
    A large area of research focuses on this problem of creating such relevant abstract spaces which are both: sufficient to solve the underlying task; and which possess simpler dynamics to learn and plan upon \citep{tamar2016value,silver2017predictron,farahmand2017value,farahmand2018iterative,farquhar2017treeqn,oh2017value,schrittwieser2020mastering,hafner2020mastering,ye2021mastering}. In particular, the work of \citet{hafner2020mastering} is interesting since it produces discrete abstract states such as used in our solution.
    
    
    


\bibliographystyle{unsrtnat}
\bibliography{biblio}

\newpage
\appendix
\section{Hyperparameters and computational resources}\label{sec:hyper}
    We refer to the reader to Figure \ref{fig:model_arch} for the model architecture. We describe in Table \ref{hyperparameter-arch-table} the hyperparameters used with this architecture. The size of the dataset of trajectories is chosen to be sufficiently large to avoid overfitting.
    
    We train the model with $2$ days of compute on a \texttt{NVIDIA RTX 6000} with \texttt{amp} on. Generating the dataset offline takes $60$ cpu hours. Running the tree search with $1250$ expansions for 250 episodes (each of $200$ steps) takes $3$ hours without \texttt{amp}.
    
    \begin{table}[h]
    \caption{List of hyperparameters for the model's architecture and its training.}
    \label{hyperparameter-arch-table}
    \begin{center}
    \begin{tabular}{ll}
        \multicolumn{1}{c}{\bf Parameter}  &\multicolumn{1}{c}{\bf Value/method}
        \\ \hline \\
        \#dimensions at the input/out of a transformer layer    & $256$ \\
        \#heads for attention & $4$ \\
        \#transformer layers & $3$ \\
        \#dimensions of the layer in the mlp of transformer layer & $256$\\
        dropout in transformer & $0.1$ \\
        activations & relu \\
        \#layers in mlps at transformer encoder output & 1\\
        \#dimensions of the layers in mlps at transformer output & 512\\
        \#dimensions of the hidden state in GRU & 32\\
        optimizer & Adam\\
        learning rate & 1e-4\\
        batch size & 250\\
        \#trajectories in dataset (each of 200 steps) & $1e6$\\
        training time (observed number of optimization steps for convergence) & $750\times 1e3$
    \end{tabular}
    \end{center}
    \end{table}
    
    All the results using tree-search presented in the paper uses the hyperparameters in Table \ref{hyperparameter-mcts-table}. These hyperparameters are found using Bayesian optimization implemented by \citet{head_tim_2020_4014775} for the case with 1.250 expansions in the tree.
    \begin{table}[h]
    \caption{List of hyperparameters for the tree search algorithm.}
    \label{hyperparameter-mcts-table}
    \begin{center}
    \begin{tabular}{ll}
        \multicolumn{1}{c}{\bf Parameter}  &\multicolumn{1}{c}{\bf Value}
        \\ \hline \\
        Temperature ($T$) & $0.55$\\
        $C_1$ & $0.57$\\
        $C_2$ & $16.15$\\
        Stochastic branching factor & $3$\\
        $\gamma$ & 0.99
    \end{tabular}
    \end{center}
    \end{table}
    
    \begin{table}[h]
    \caption{From \cite{wang2021alchemy}, list of publicly available hyperparameters for the V-MPO agent. The architecture is defined in \cite{parisotto2020stabilizing} and the training algorithm in \cite{song2019v}.}
    \begin{center}
    \begin{tabular}{ll}
        \multicolumn{1}{c}{\bf Parameter}  &\multicolumn{1}{c}{\bf Value/method}
        \\ \hline \\
        agent discount & $0.99$ \\
        TrXL MLP size & $256$ \\
        TrXL number of layers & $6$ \\
        TrXL number of heads & $8$\\
        TrXL Key/Value size & $32$ \\
        $\epsilon_{\eta}$ & $0.5$ \\
        $\epsilon_{\alpha}$ & $0.001$\\
        $T_{\text{target}}$ & 100\\
        $\beta_{\pi}$ & $1.0$\\
        $\beta_{V}$ & $1.0$\\
        \#episodes used & $1e9$
    \end{tabular}
    \end{center}
    \end{table}

\section{Model architecture}\label{sec:model-details}
    \begin{figure}[ht]
            \begin{center}
            \includegraphics[width=0.75\linewidth]{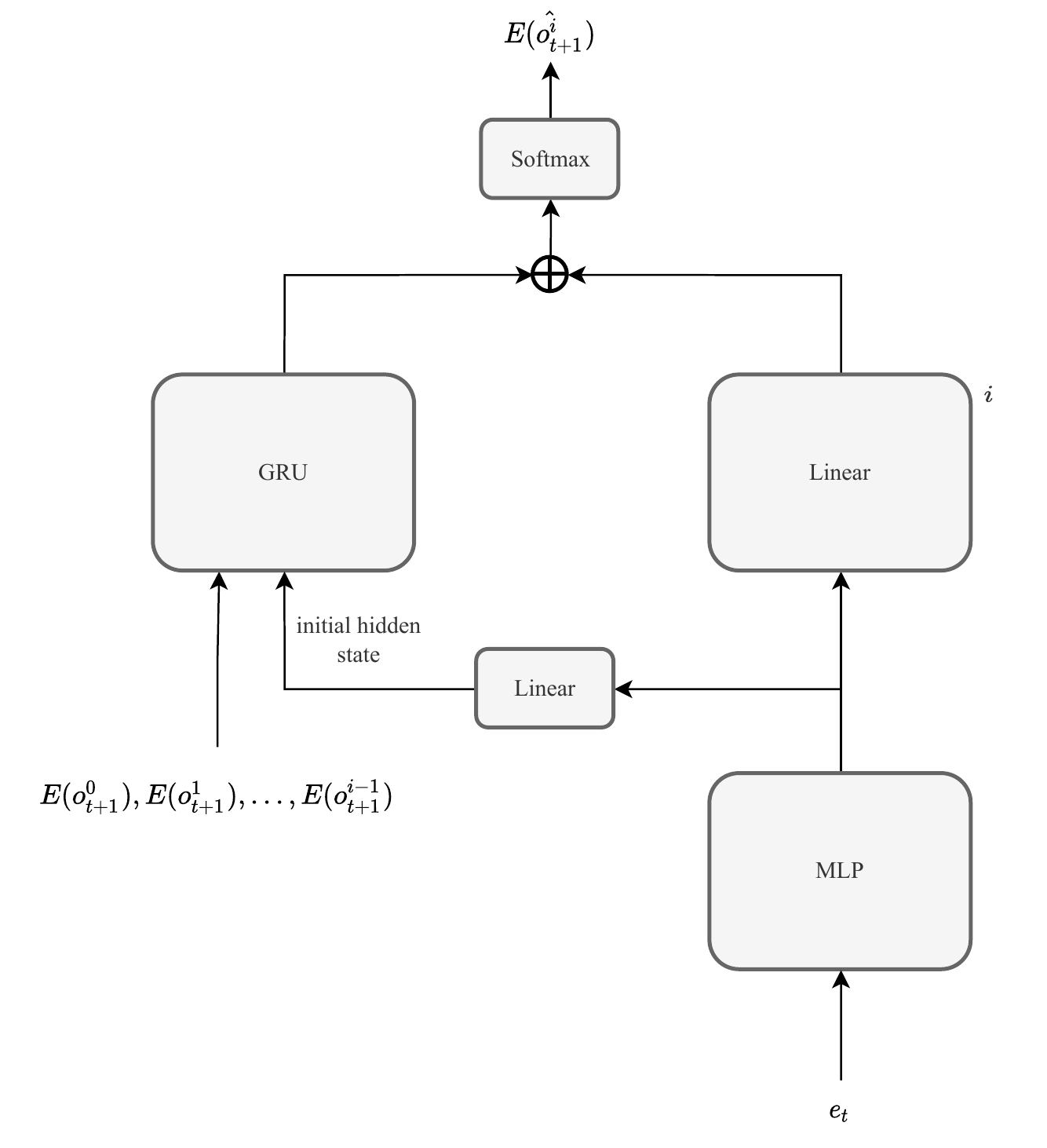}
            \caption{A more detailed view of the prediction head at training time using teacher forcing when predicting dimension $i$ of the observation at timestep $t$. Each dimension of the observation space as its own linear head (on the right in the Figure), so we take the corresponding $i$th component.}
            \label{fig:teacher-forcing}
            \end{center}
        \end{figure}

\end{document}